\documentclass[letterpaper, 10 pt, conference]{IEEEtran}  

\IEEEoverridecommandlockouts                                

\usepackage{amsmath, amssymb} 
\usepackage[english,dutch]{babel}
\usepackage{graphicx}
\graphicspath{{figures/}}
\usepackage[font=footnotesize,labelfont=bf]{caption}
\usepackage{bbold} 
\usepackage{svg}
\usepackage{comment}
\usepackage{xcolor}
\usepackage{import}
\usepackage{multirow}
\usepackage{microtype}
\usepackage[
giveninits=true,
maxbibnames=99,
backend=biber,
style=ieee,
sorting=none
]{biblatex}
\usepackage{tikz}
\usepackage{textcomp}
\usepackage{hyperref}
\usepackage{lipsum}

\newcommand\copyrighttext{%
  \footnotesize \textcopyright 2024 IEEE. Personal use of this material is permitted.
  Permission from IEEE must be obtained for all other uses, in any current or future
  media, including reprinting/republishing this material for advertising or promotional
  purposes, creating new collective works, for resale or redistribution to servers or
  lists, or reuse of any copyrighted component of this work in other works.}
\newcommand\copyrightnotice{%
\begin{tikzpicture}[remember picture,overlay]
\node[anchor=south,yshift=10pt] at (current page.south) {\fbox{\parbox{\dimexpr\textwidth-\fboxsep-\fboxrule\relax}{\copyrighttext}}};
\end{tikzpicture}%
}

\title{\LARGE \bf
Identification and validation of the dynamic model of a tendon-driven anthropomorphic finger
}

\author{
Junnan Li$^{1,\star }$, Lingyun Chen$^{1,2,\star }$, Johannes Ringwald$^{1}$, Edmundo Pozo Fortuni\'c$^{1}$,\\ Amartya Ganguly$^{1}$, and Sami Haddadin$^{1,2}$%
\thanks{$^{1}$The authors are with Munich Institute of Robotics \& Machine Intelligence, Technische Universität München (TUM), Germany $^2$ and also with the Centre for Tactile Internet with Human-in-the-Loop (CeTI). 
This work was supported by the Federal Ministry of Education and Research of the Federal Republic of Germany (BMBF) by funding the project AI.D under Project no. 16ME0539K and the German Research Foundation (DFG, Deutsche Forschungsgemeinschaft) as part of Germany’s Excellence Strategy – EXC 2050/1 – Project ID 390696704 – Cluster of Excellence “Centre for Tactile Internet with Human-in-the-Loop” (CeTI) of Technische Universität Dresden. Email:\texttt{\{junnan.li, lingyun.chen, johannes.ringwald, edmundo.pozo, amartya.ganguly, haddadin\}@tum.de}.
}
\thanks{$^{*}$The first two authors contributed equally to this work.}
}

\providecommand  {\review}     [1]{{\color{review}#1}}

\colorlet{ag}{magenta}
\colorlet{lc}{orange}
\colorlet{ljn}{blue}
\colorlet{epf}{red}
\colorlet{review}{black}

\newcommand{\test}[1][]{%
\ifthenelse{\equal{#1}{}}{omitted}{given}%
}

\global\long\def\mymatrix #1{\boldsymbol{#1}}%
\global\long\def\myvec #1{\boldsymbol{#1}}%

\global\long\def\realset{\ensuremath{\mathbb{R}}}%

\renewcommand{\baselinestretch}{0.96}

\begin{document}

\maketitle
\copyrightnotice




\begin{abstract}
This study addresses the absence of an identification framework to quantify a comprehensive dynamic model of human and anthropomorphic tendon-driven fingers, which is necessary to investigate the physiological properties of human fingers and improve the control of robotic hands. 
First, a generalized dynamic model was formulated, which takes into account the inherent properties of such a mechanical system. This includes rigid-body dynamics, coupling matrix, joint viscoelasticity, and tendon friction. 
Then, we propose a methodology comprising a series of experiments, for step-wise identification and validation of this dynamic model. 
Moreover, an experimental setup was designed and constructed that features actuation modules and peripheral sensors to facilitate the identification process.
To verify the proposed methodology, a 3D-printed robotic finger based on the index finger design of the Dexmart hand was developed, and the proposed experiments were executed to identify and validate its dynamic model. 
This study could be extended to explore the identification of cadaver hands, aiming for a consistent dataset from a single cadaver specimen to improve the development of musculoskeletal hand models.
\end{abstract}

\section{INTRODUCTION}

%
%





Much effort has been devoted to exploring and modelling the physiological properties of human hands in order to improve the understanding of human hand functions and human motor control \cite{mcfarland2022musculoskeletal, engelhardt2020new, ma2019validated, mirakhorlo2018musculoskeletal, barry2018development}. 
For example, the study \cite{ma2019validated} proposed a kinematic model with respect to the tendon paths using obstacle set algorithms and validated the moment arm values with the cadaver data. 
An open-source musculoskeletal hand-wrist model is proposed in \cite{mcfarland2022musculoskeletal} incorporating the properties of moment arms and passive joint torques, and is validated within multiple scenarios, such as lateral pinch and passive grasp/release.  
Meanwhile, anthropomorphic robotic tendon-driven hands have been developed as the counterpart of the human hand, aiming to achieve the dexterity of human hands in daily tasks \cite{grebenstein2012hand, palli2014dexmart, zhu2022anthropomorphic} and to provide a platform for exploring the biomechanical properties of human hands by mimicking the geometry and tendon structure of human hands \cite{deshpande2011mechanisms}\cite{xu2016design}. 



There are challenges that remain in modelling the properties of both, tendon-driven robotic as well as human fingers and hands, such as the nonlinear moment arm, the passive joint viscoelasticity, and tendon transmission friction, which play crucial roles in finger dynamics.
Extensive experiments have been designed to quantify these parameters in musculoskeletal finger models through cadaver and in-vivo studies \cite{an1983tendon, francis2021moment, valero2000quantification, lu2018novel, deshpande2011contributions, }. 
For example, the tendon excursion method was proposed and applied to identify the moment arm curves of cadaver index fingers in \cite{an1983tendon} and was extended to all fingers in study \cite{francis2021moment}. The study \cite{deshpande2010acquiring} designed an experiment using a regression method to determine the moment arm curves of tendons using the ACT test-bed \cite{deshpande2011mechanisms}.
The passive viscoelasticity property of the human index finger has been investigated in \cite{li2006robot}\cite{deshpande2011contributions}\cite{kuo2012muscle} through in vivo experiments, in which the study \cite{deshpande2011contributions} indicated that the joint viscoelasticity dominates the finger dynamics over rigid body dynamics during movement. The fingertip forces as a result of tendon forces were explored in cadaver studies \cite{valero2000quantification}\cite{lu2018novel}, revealing characteristics of force transmission through the biological system. The mapping from tendon forces to the fingertip forces is also taken as a criterion for validating the musculoskeletal hand models \cite{mcfarland2022musculoskeletal}. 
On the other hand, \cite{palli2011modeling} devoted efforts to describe the dynamic friction phenomenon in the tendon transmission system of robotic fingers using a LuGre-like model on a standalone experimental setup. The study \cite{lange2021friction} proposed a method identifying the friction model of the finger on the Awiwi hand finger without disassembling the finger. 

However, the main barrier to biomechanical finger modelling is the limited amount of experimental data \cite{mcfarland2022musculoskeletal}. 
Moreover, most cadaver studies aim at only one aspect of the physiological properties at a time, i.e. moment arm curves, passive joint properties, and fingertip forces, resulting in the lack of an identification pipeline that quantifies and validates a complete finger dynamic model from a consistent dataset.
Having a consistent musculoskeletal hand model is necessary to simulate human hand functions as the study shows that the models derived from different datasets may lead to a distinction up to 180\% in muscle force estimation \cite{goislard2018importance}. 

\begin{figure*}[tb]
    \centering
    \fontsize{7pt}{7pt}\selectfont
    \def\svgwidth{0.98\textwidth}
    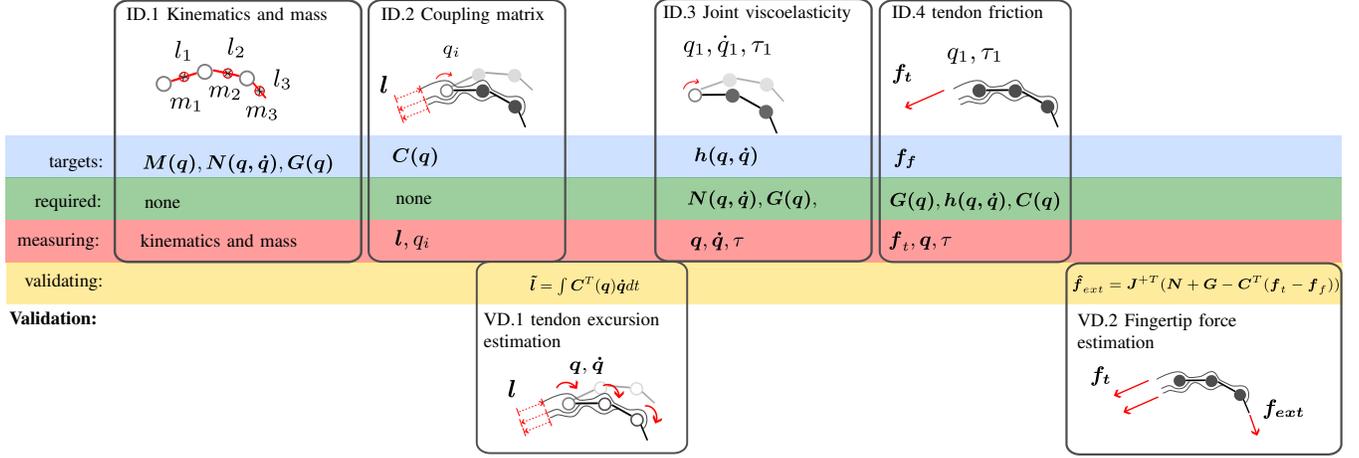
    \caption{The identification and validation methodology overview: identification (ID) and validation (VD) experiments in sequence targeting different parameters. The tendon-driven finger is depicted in grey, in which the solid and hollow circles mean the free-to-move and fixed joints, respectively. The parameters to be identified or validated in each experiment are formulated in blue rows. The necessary parameters to identify the target variables in the model are listed in the green row. The measurements in each experiment are highlighted in red. The mathematical equations of validation scenarios are summarized in the yellow row. }
    \label{fig: method ID flowchart}
    \vspace{-15pt}
\end{figure*}

Therefore, we strive to address the challenge of the lack of a comprehensive identification pipeline that allows to quantify and validate a musculoskeletal finger model on a single cadaver specimen.  
We first formulate a general dynamic model for tendon-driven fingers, incorporating the critical properties of both human and robotic fingers. 
Then, a methodology comprising sequential identification and validation experiments is proposed, which provides a pipeline to step-wise derive and evaluate all parameters in the dynamic model via a consistent dataset from one finger.
The associated identification test-bed is developed as a portable and extendable platform to conduct the experiments and equipped with necessary components, i.e. actuation module and sensors, to conduct the proposed experiments. 
As a preliminary validation of the methodology before any cadaver study, we design an anthropomorphic tendon-driven finger based on the Dexmart hand design \cite{borghesan2010design} and identify the dynamic model of this particular finger.

\section{METHODS}

\subsection{Dynamic Modelling}

\subsubsection{General equation}
The dynamics of A tendon-driven finger with $m$ joints and $n$ tendons can be formulated as 
\begin{align}
\mymatrix M( \myvec q) \myvec {\ddot{ \myvec q}}  + \mymatrix N( \myvec q, \myvec {\dot{q}}) & + 
\mymatrix G(\myvec q) + \myvec h ( \myvec q, \myvec {\dot{q}}) =  \notag \\
\mymatrix C^T (\myvec q)  (\myvec f_t - \myvec f_f (\myvec f_t,\myvec {\dot{l}}) ) &+ \mymatrix J_{ext}^{{T}} \myvec {f}_{ext}; 
\label{eq: general dynamic formula 1}  
\end{align}

where, $\mymatrix M \in \realset^{m \times m}$, $\mymatrix N \in \realset^{m} $, $\mymatrix G \in \realset^{m}  $ are the inertia matrix, centrifugal and Coriolis vector, and gravity vector of the finger, respectively.
The external force is denoted as $\myvec {f}_{ext}$ and its associated Jacobian matrix $\mymatrix J_{ext}$. 
The direction of joint flexion or abduction is defined as positive.
The remaining terms in the dynamic model are explained in the following subsections. 

\subsubsection{Coupling matrix}
The coupling matrix $\mymatrix C^T(\myvec q) \in \realset^{m \times n}$, also known as the moment arm matrix, reveals the transmission relationship between the tendon force $\myvec f_t \in \realset^{n}$ and the joint torque $\myvec \tau \in \realset^{m}$.
In a human hand, the muscle forces are transmitted through tendons along specific physiological paths constrained by the tendon via points, bone geometry, and soft tissues such as ligaments and the tendon pulley. Moreover, the complex interconnected tendon structure, i.e. the extensor mechanism, leads to the non-linearity of the moment arm property compared to mechanical tendon guiding. 
The transpose of the coupling matrix, $\mymatrix C(\myvec q)$, represents the kinematic relationship between the tendon excursion velocity $\myvec {\dot{l}}$ and the joint velocity $\myvec {\dot{q}}$:
\begin{equation}
\myvec l = \myvec f(\myvec q),  \quad \myvec {\dot{l}} = \mymatrix C (\myvec q) \myvec {\dot{q}},
\label{eq: moment arm matrix}
\end{equation}


\subsubsection{Joint Viscoelasticity}
The vector $\myvec h(\myvec q, \myvec{\dot{q}})$ is formed by the joint pose and velocity-dependent torque, including the joint viscoelasticity or friction torques. 
In terms of the human finger, joint viscoelasticity is a critical property that contributes to the system dynamics over mass and inertia terms \cite{deshpande2011contributions}\cite{kuo2012muscle}.
The joint viscoelasticity of the human finger is derived from the stiffness of the muscle-tendon unit, joint capsules, and surrounding tissues. Therefore, it performs non-linearity to the joint states $ \myvec q$ and $ \myvec {\dot{q}}$.
On the other hand, $\myvec h(\myvec q, \myvec{\dot{q}})$ torque in a robotic finger is embodied as, for example, the elastic torque due to the embedded spring-like elements at the joints and the joint friction.
It is noted that the joint friction is solely joint-state-dependent and is distinguished from the tendon transmission friction $ \myvec f_f$, which is related to tendon forces. 

Therefore, $\myvec h(\myvec q, \myvec{\dot{q}})$ can be formulated as the combination of joint elastic torque $\myvec \tau_k (\myvec q)$, joint damping $\myvec \tau_d (\myvec {\dot{q}})$, and joint friction $\myvec \tau_{f,q} (\myvec q, \myvec {\dot{q}}))$ torques:
\begin{equation}
    \myvec h(\myvec q, \myvec {\dot{q}}) = \myvec \tau_k (\myvec q) + \myvec \tau_d (\myvec {\dot{q}}) + \myvec \tau_{f} (\myvec q, \myvec {\dot{q}}).
\label{eq: h term}
\end{equation}

\subsubsection{Friction of tendon transmission system}
The force vector $\myvec f_f (\myvec f_t,\myvec {\dot{l}})$ represents the tendon friction along the tendon transmission, which has a major influence in the robotic finger system \cite{palli2011modeling}\cite{lange2021friction}.
Various models can be used to describe the reduction of tendon forces due to friction, such as Coulomb, Stribeck, viscous, and LuGre dynamic friction models \cite{palli2011modeling}. In this paper, we represent tendon friction $\myvec f_f (\myvec f_t,\myvec {\dot{l}})$ as a function of the tendon force $\myvec f_t $ and tendon velocity $\myvec {\dot{l}}$, 



\subsection{Identification Approach}

We propose a unified approach that includes a sequence of identification experiments to systematically identify parameters in the dynamic model, followed by validation scenarios to assess the accuracy of the model parameters. For clarity, \textit{ID} and \textit{VD} are abbreviations for \textit{Identification} and \textit{Validation}, respectively.

\subsubsection{ID.1: kinematic and inertia parameters}
The kinematic and inertia parameters are necessary to calculate $\mymatrix M, \mymatrix N,\mymatrix G$ terms in \eqref{eq: general dynamic formula 1} as the rigid body dynamics of the robotic and anatomical fingers. 
As a robotic finger, these parameters can be extracted from the CAD design as well as the material information. 
Conversely, kinematic parameters of human hands can be measured geometrically \cite{mirakhorlo2016anatomical}, while bone anatomical parameters are inferred from bone length and known bone density \cite{mirakhorlo2016anatomical}\cite{goislard2018importance}.

In the first step of the identification approach, we derive the kinematic and inertia parameters by measuring the geometry of the finger and estimating the mass property according to the material density, as shown in Fig.~\ref{fig: method ID flowchart} ID.1, thus the terms of $\mymatrix M(\myvec q), \myvec N(\myvec q, \myvec {\dot{q}}), \myvec G(\myvec q) $ in \eqref{eq: general dynamic formula 1} can be calculated with the given joint states $\myvec q, \myvec {\dot{q}}$. 

\subsubsection{ID.2: Coupling matrix}


According to the characteristic of coupling matrix $ \mymatrix C$ shown in \eqref{eq: moment arm matrix}, it can be derived by recording the kinematic relationship between the joint position and the tendon excursion and calculating its Jacobian.

At the beginning of the experiment, all the tendons are pre-tensioned with a small force, for example, 2 N, to avoid tendon slacking. 
The mapping $\myvec f$ in Eq.~\eqref{eq: moment arm matrix} at each joint can be measured independently by fixing all other joint positions.
The operator manually moves the target joint $q_i$ in low joint velocity mode, reaching all positions within its range of motion. The joint position and corresponding tendon excursion of all associated tendons are recorded during the movement. 
Then, the recorded mapping $f_{i,k}$ between the target joint $q_i$ and the tendon excursion $l_k$ can be fit, for instance, with a 4th-order polynomial whose derivative is considered as the moment arm $c_{i,k} (q_i)$ in the coupling matrix $\mymatrix C$. By repeating the experiment on all joints, the coupling matrix $\mymatrix C(\myvec q)$ can be derived with the given joint pose $\myvec q$:
\begin{equation}
c_{i,k} = \frac{\partial f_{i,k} }{\partial q_i}, \quad \mymatrix C(\myvec q) = \frac{\partial \myvec f }{\partial \myvec q}.
\label{eq: moment arm elements}
\end{equation}


\subsubsection{VD 1: Tendon excursion estimation}

To validate the coupling matrix $\mymatrix C(\myvec q)$ derived from the ID.1, A kinematic scenario was designed by comparing the estimated and measured tendon excursion while moving the finger joints.

All tendons of the finger were in pre-tension to avoid tendon slacking. All finger joints remained free to move. The operator manually held the fingertip and moved the finger gently with a low velocity. The joint position and velocity were recorded to calculate the coupling matrix and to estimate the tendon excursion as
\begin{equation}
\hat{\myvec l} = \int \mymatrix C (\myvec q) \myvec {\dot{q}} \mathrm{d}t.
\label{eq: VD1: tendon excursion}
\end{equation}
By comparing the estimated tendon excursion $\hat{\myvec l}$ with the measured $\myvec l_m$, the parameters of the coupling matrix were evaluated.

\subsubsection{ID.3: Joint viscoelasticity torque}

In the experiment of determining the joint viscoelasticity torque, the finger remains passive, and tendons were in slack to avoid tendon-related torque. In this case, the dynamic model in \eqref{eq: general dynamic formula 1} is simplified to
\begin{equation} 
\mymatrix N( \myvec q, \myvec {\dot{q}}) + 
\mymatrix G(\myvec q) + \myvec h ( \myvec q, \myvec {\dot{q}}) = \myvec {\tau}_{ext}; 
\label{eq: ID.3 joint viscoelasticity}
\end{equation}
As $\mymatrix N, \mymatrix G$ have been identified in ID.1, $ \myvec h( \myvec q, \myvec {\dot{q}})$ can be derived if the external torque $\myvec {\tau}_{ext}$ is measurable. 
To this aim, an external motor equipped with a torque sensor was applied to drive the target joint, such that, the applied torque could be measured, which is described in Sec.\ref{sec: Experimental setup}. 
For a robotic finger, the torque sensor shaft was connected to the joint shaft using a shaft coupler. 
Each joint was identified individually with the other joints fixed during the experiment.

The experiment is divided into two phases: static and dynamic, investigating the effects of joint position and velocity, respectively. 
In the static phase, the joint elastic $\myvec {\tau}_k (\myvec q) $ and static friction torque $\myvec {\tau}_f (\myvec q) $ were determined. The target joint started from a complete extension position and was controlled to stop at sampled joint positions, for example, every 10$^{\circ}$, while finishing at maximum flexion. 
The joint was then driven back from flexion to extension while stopping the joint at the same sampled positions. 
When the joint was close to stable at each sampled position, i.e. low joint velocity to ensure the moving direction, the external torque and joint position were measured. 
The gravity torque $\mymatrix G(\myvec q)$ was calculated with the recorded joint position. Thus, the $\myvec {\tau}_k(\myvec q)$ and static part of $ \myvec {\tau}_f$ in Eq. \eqref{eq: h term} can be represented by 2nd-order polynomials.

In the dynamic phase, the entire joint viscoelasticity torque $\myvec h(\myvec q, \myvec{\dot{q}})$, including elastic and viscous friction torques, can be identified with a similar procedure by driving the target joint at different joint velocities within the range of joint motion. 
Thus, the joint elastic and static torque derived from the static phase was augmented by the joint damping $\myvec {\tau_d}$ and viscous friction.


\subsubsection{ID.4: Tendon transmission friction}

\begin{figure*}[tb]
    \centering
    \includegraphics[width=0.98\textwidth]{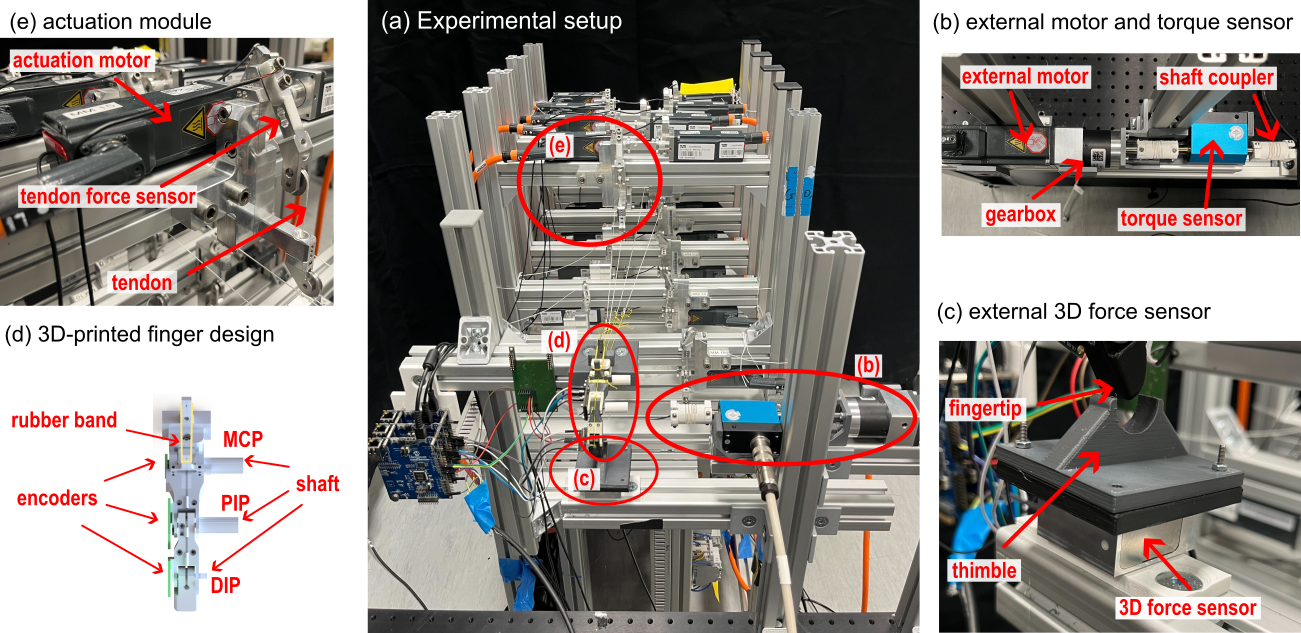}
    \caption{The overview of the identification test-bed and robotic finger. The external motor with torque sensor is shown in (b). The external 3-dimensional force sensor with the thimble is depicted in (c) for measuring the fingertip force. \review{Fig.} (d) illustrates the finger design based on the Dexmart hand with joint shafts and encoders. A rubber band is integrated as a stiffness component at the MCP joint. One of the eighteen actuation modules with a customized pulley mechanism and tendon force sensor is shown in (e). }
    \label{fig: method experiment setup}
    \vspace{-15pt}
\end{figure*}

This experiment is designed to identify the reduction of tendon forces due to the frictional attachments through transmission.
We assume the tendon friction $\myvec f_f (\myvec f_t,\myvec {\dot{l}})$ is static and depends on the input tendon force $\myvec f_t$ and the direction of tendon excursion $ \mathrm{sign}( \myvec {\dot{l}})$.
Thus, we design a static scenario to explore the mapping of tendon force friction to the tendon force.
Each tendon is identified individually, while the other tendons remain slack.
Before the experiment, the finger is posed in a neutral configuration, and the most distal joint connected by the target tendon is linked to the external torque sensor with a shaft coupler.
Then, the tendon was pulled with a discrete level of tension by the identification test-bed, which is described in Sec.\ref{sec: Experimental setup}. The resultant joint torque was measured by the torque sensor. The model was simplified for this scenario to 
\begin{equation} 
\mymatrix G(\myvec q) + \myvec h ( \myvec q, \myvec {\dot{q}}) = \mymatrix C^T (\myvec q)  (\myvec f_t - \myvec f_f (\myvec f_t,\mathrm{sign}( \myvec {\dot{l}}))) + \myvec {\tau}_{ext}, 
\label{eq: ID.4 tendon friction}  
\end{equation}
where $\mymatrix G(\myvec q)$, $\myvec h ( \myvec q, \myvec {\dot{q}})$, and $\mymatrix C^T$ were derived by the recorded joint position $\myvec q$. Thus, the reduction of tendon force $\myvec f_f$ was calculated with the measured tendon force $\myvec f_t$ and external torque $\myvec {\tau}_{ext}$ necessary to maintain the finger pose. 


\subsubsection{VD.2: Fingertip force estimation}

Calculating the static fingertip force, corresponding to zero joint velocity, entails understanding the coupling matrix $\mymatrix C$, finger gravity torque $\mymatrix G$, joint viscoelasticity torque $\myvec h$, and tendon friction $ \myvec f_f (\myvec f_t,\mathrm{sign}( \myvec {\dot{l}}))$ as outlined:
\begin{equation} 
\mymatrix G(\myvec q) + \myvec h ( \myvec q, \myvec {\dot{q}}) = 
\mymatrix C^T (\myvec q)  (\myvec f_t - \myvec f_f (\myvec f_t,\mathrm{sign}( \myvec {\dot{l}}))) + \mymatrix J_{ext}^{{T}} \myvec {f}_{ext}.
\label{eq: VD2: Fingertip force}
\end{equation}
Similar to the experiment setup proposed in \cite{valero2000quantification}, we place the finger in a neutral pose. 
The position of the external 3D force sensor was adjusted such that the fingertip was slightly attached to the surface of the sphere interface. 
We individually controlled each tendon to apply discrete levels of forces from 0--15 N while all other tendons remained at zero tension. 
The tendon tension and the resultant fingertip forces were recorded by an external force sensor. 
The fingertip force was estimated by the identified model in Eq.~\eqref{eq: VD2: Fingertip force}.
The model accuracy was validated by comparing the measured and the estimated forces.




\subsection{Experimental Setup}
\label{sec: Experimental setup}

\subsubsection{Finger design}

The targeted finger was custom-designed based on the Dexmart hand tendon configuration \cite{borghesan2010design, palli2014dexmart} with modifications on the structure to increase the manipulation complexity and host additional sensors and stiffness components. The detailed finger design is shown in Fig.~\ref{fig: method experiment setup} (d). Tendon configuration spec detail can also go here from the description in Sec.~\ref{sec: result ID.1}. 

The absolute position encoders (AA5048B, ams-OSRAM AG) and their corresponding magnet were located at each finger joint axis. A custom-made EtherCAT slave PCB was used to transfer the joint position data to the main system bus at a 1 kHz rate using an ARM-Cortex M4F microcontroller and a LAN9253 controller (Microchip).


Additional structures, including a rubber band holder and shaft extender for each axis were attached to host additional stiffness elements and connections to the external torque sensor, respectively. This allows an easy modification of the joint stiffness by changing the rubber band.

\subsubsection{Identification test-bed}
This is a modular, extendable and portable test-bed design for identifying and manipulating both tendon-driven robotic and human hands, as shown in Fig.~\ref{fig: method experiment setup}(a). 
Eighteen (18) actuation modules were implemented, which are able to connect the majority of the extrinsic muscles of the human hand. 
Each actuation module is comprised of a permanent magnet synchronous motor (AM8112-1F10, Beckhoff Automation) with an embedded position sensor and its corresponding closed-loop servo controller (EL7211-0010, Beckhoff Automation), a pulley-based guiding system mechanism with an in-line one-dimensional force sensor (KD60 100 N, ME Messsysteme), a corresponding GSV-8 differential amplifier and an Analog-to-Digital-Converter (EL3008, Beckhoff Automation) for measuring the corresponding tendon force, as shown in Fig.~\ref{fig: method experiment setup}(e). 
All controllers and ADCs are connected to the master control PC via the EtherCAT bus, providing a 1 kHz sample rate.
One floating 3D force sensor (K3D40 $\pm$50 N, ME Messsysteme) in Fig.~\ref{fig: method experiment setup}(c) was integrated into the system, whose position was adapted to the finger pose such that the fingertip force is measured.
Additionally, an external motor, as shown in Fig.~\ref{fig: method experiment setup}(b), was implemented with a torque sensor as an extra finger joint actuator, with its position adaptable to finger poses, to actuate the finger with a measurable torque. This actuator was used in the identification experiments, i.e., ID.3 and ID.4, to move and stabilize the joint at desired positions and speeds.
All data acquisition, processing, and motor control were implemented in Matlab/Simulink 2018b. The customized Matlab toolbox was used for dynamic modelling \cite{li2024fingertip}.


\section{RESULTS}

\subsection{ID.1 Kinematic and mass parameters}
\label{sec: result ID.1}

\begin{figure}[tb]
    \centering
    \includegraphics[width=0.48\textwidth]{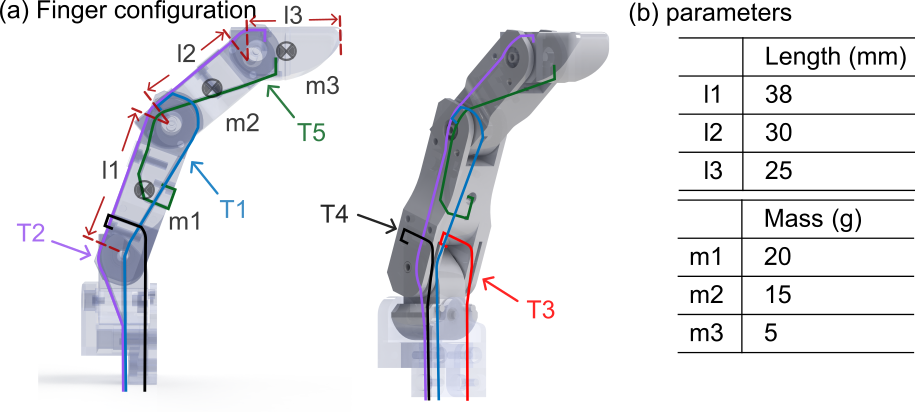}
    \caption{The illustration of the finger with the tendon configuration and associated notation. The kinematic and estimated mass parameters are depicted in (b).}
    \label{fig: result ID kinematics}
    \vspace{-10pt}
\end{figure}

In Fig.~\ref{fig: result ID kinematics}(a), the tendon configuration of the finger and associated notation from T1 to T5 are depicted. 
T1 to T4 are active tendons, while T5 is a passive tendon coupled the PIP joint with the DIP joint. The MCP abduction joint was fixed during the experiments. The kinematic parameters are derived from the CAD file. The mass of each link is estimated according to the weight of 3D-printed materials, screws and attached PCB board as shown in Fig.~\ref{fig: method experiment setup}(d).  

\subsection{ID.2 Coupling matrix}

\begin{figure}[tb]
    \centering
    \fontsize{6pt}{6pt}\selectfont
    \def\svgwidth{0.48\textwidth}
    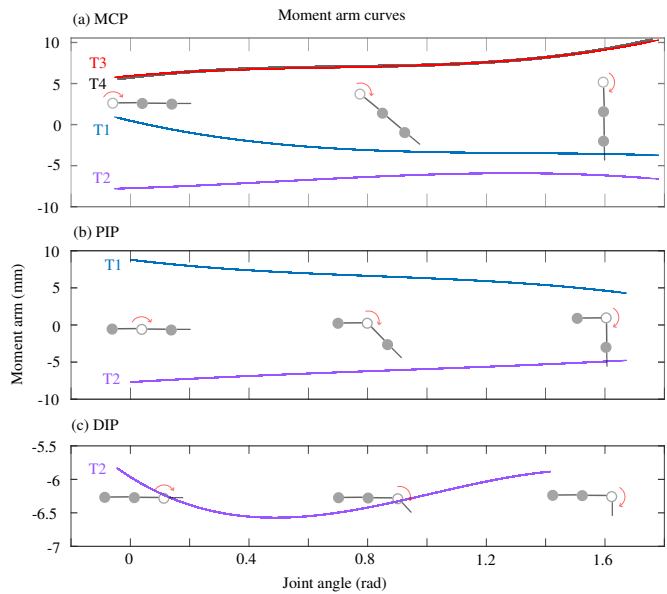
    \caption{The moment arm curves from ID.1 experiment. The curves of all four active tendons at associated joints are visualized. The positive and negative moment arm values refer to the directions of flexion and extension, respectively. }
    \label{fig: result ID coupling}
    \vspace{-1pt}
\end{figure}

\begin{table}[tb]
\centering
\caption{The range of moment arms in mm (max; min)}
\setlength{\tabcolsep}{8pt}
\renewcommand{\arraystretch}{1.2}
\begin{tabular}{|c|c|c|c|c|}
\hline
    & T1       & T2        & T3       & T4       \\ \hline
MCP & 3.7; -0.9 & -5.9; -7.7 & 10.3; 5.8 & 10.5; 5.5 \\ \hline
PIP & 8.8; 4.3  & -4.8; -7.7 &   --      &   --      \\ \hline
DIP & --  & -5.8; -6.6 &    --        &   --       \\ \hline
\end{tabular}
\label{tab: result ID coupling}
\vspace{-10pt}
\end{table}

\begin{figure}[tb]
    \centering
    \fontsize{6pt}{6pt}\selectfont
    \def\svgwidth{0.49\textwidth}
    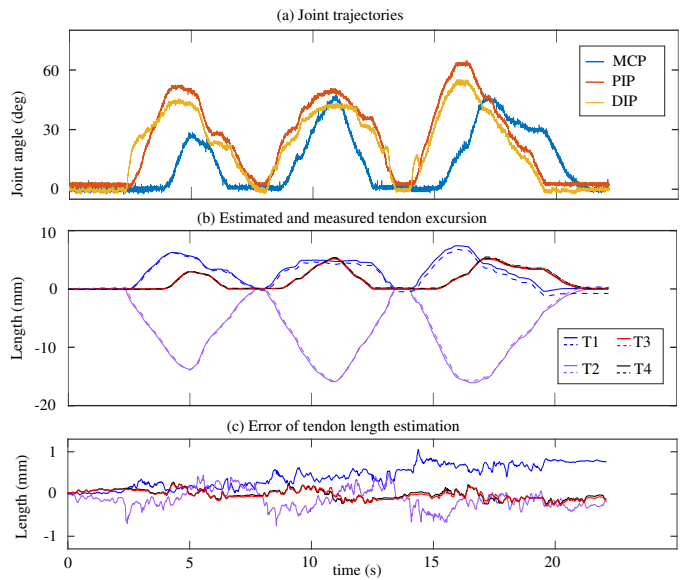
    \caption{The results of the VD.1 experiment. The joint trajectories of MCP, PIP, and DIP are visualised in (a). All four active tendon excursions are estimated online (dashed lines) and compared to the measurements (solid lines). The errors between the estimation and measurement of tendon length are depicted in (c).}
    \label{fig: result VD coupling}
    \vspace{-15pt}
\end{figure}

The moment arm curves of all active tendons at associated joints are visualised in Fig.~\ref{fig: result ID coupling}. The ranges of moment arm values for all tendons are summarized in Tab.\ref{tab: result ID coupling}. The moment arm curves of T3 and T4, shown in red and black in Fig.~\ref{fig: result ID coupling}(a), are similar at the MCP joint varying from 5.8 to 10.3 mm.
The range of variation matches the finger design, where both tendons are flexors at the MCP joint and also have symmetric tendon paths. 
When the MCP joint is close to 90$^{\circ}$, the moment arms of T3 and T4 are close to 10.3 mm, which is larger than the radius of the pulley. It can be explained that the tendons lose their attachment to the pulley; thus, the distance of the joint axis to the tendon pathway is increased.  
The moment arm value of T1 (blue) at the MCP joint decreases from 1 to -3.7 mm, indicating that the function of T1 changes from a flexor to an extensor while the joint angle increases. The curve of T2 fluctuates between -5.8 to -6.5 mm at the DIP joint, depicted in Fig.~\ref{fig: result ID coupling}(c), since the contact surface is not concentric with the DIP joint axis.

\subsection{VD.1 Tendon excursion estimation}

\review{Fig.} \ref{fig: result VD coupling} shows the results of the VD.1 experiment for validating the coupling matrix. The operator randomly manipulates the finger, with the resulting joint trajectories illustrated in Fig.~\ref{fig: result VD coupling}(a). The coupling matrix is calculated in real-time with the measured joint position. The joint velocity is derived by online differentiating the filtered joint position signals. 
The excursions of all tendons during the motion are estimated in real-time according to Eq.~\eqref{eq: VD1: tendon excursion}, shown as dashed lines in Fig.~\ref{fig: result VD coupling}(b), and compared to the measurement, depicted with solid lines. The errors between the estimation and measurement are outlined in Fig.~\ref{fig: result VD coupling}(c). 
Due to the similar tendon paths, the tendon excursions of T3 and T4 are estimated and measured analogously. The mean error between the estimations and measurements of T3 and T4 along the trajectory are 0.008 and 0.023 mm, respectively. The tracking error of T1 starts to accumulate from zero at the beginning to 0.8 mm at the end of the trajectory, with a mean error of 0.424 mm. This probably because of the measurement noise and/or online low-pass filter to derive joint velocity. 

\subsection{ID.3 Joint viscoelasticity torque}

\begin{figure}[tb]
    \centering
    \includegraphics[width=0.48\textwidth]{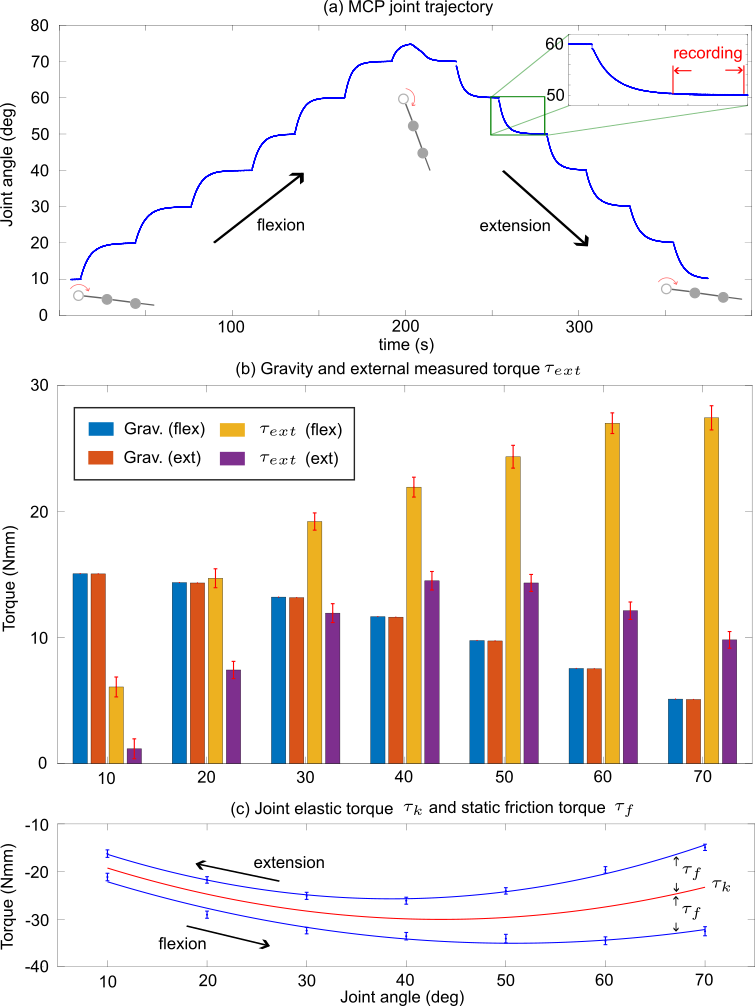}
    \caption{The result of the joint elastic and static friction torque of MCP joint in ID.3 experiment. The MCP joint was stabilized by the external motor at sampled positions while flexing and extending within 10$^{\circ}$ to 70$^{\circ}$ visualized in (a). The gravity torque in the flexion and extension phase are calculated and depicted as blue and orange bars in (b). The measured external torques are shown in yellow and purple with standard deviation for flexing and extending phases, respectively. The resultant joint elastic torque $\tau_k$ and direction-dependent static friction torque $\tau_f$ are depicted in (c).}
    \label{fig: result ID h 1}
    \vspace{-15pt}
\end{figure}

\begin{figure}[tb]
    \centering
    \fontsize{6pt}{6pt}\selectfont
    \def\svgwidth{0.48\textwidth}
    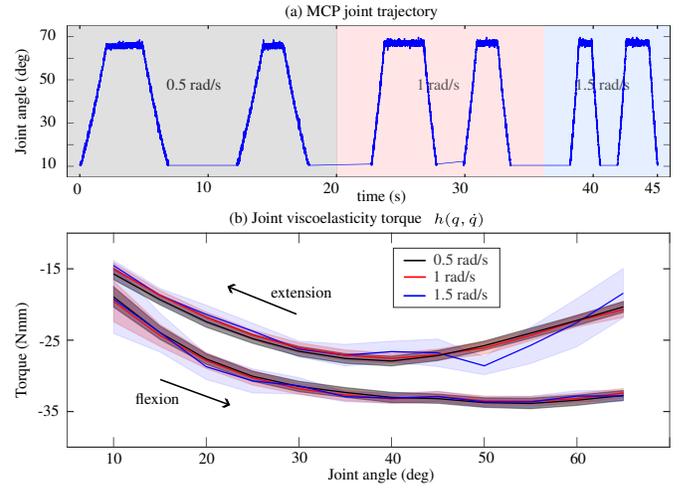
    \caption{The joint viscoelasticity torque $ h(q,\dot{q}) $ of the MCP joint with different joint velocities. The MCP joint is driven by the external motor within 10$^{\circ}$ to 65$^{\circ}$ at three different joint velocities, decoded in black, red, and blue in (a). The measured joint viscoelasticity torques according to three different joint velocities are shown against the joint angle with standard deviation in (b). }
    \label{fig: result ID h visco 1}
    \vspace{-10pt}
\end{figure}
\review{Fig.} \ref{fig: result ID h 1}-\ref{fig: result ID h visco 1} show the results of static and dynamic phases while identifying the MCP joint viscoelasticity torque. In the static experiment phase, the external motor held the MCP joint at sampled joint positions, i.e. at every 10$^{\circ}$, while moving forward (flexion) and backwards (extension) within the range of 10$^{\circ}$ to 70$^{\circ}$, as shown in Fig.~\ref{fig: result ID h 1}(a). The data was recorded while the MCP joint was close to being stable at sampled positions, i.e. joint velocity tends to zero, highlighted in red in Fig.~\ref{fig: result ID h 1}(a), to ensure the direction of joint friction. The calculated gravity torque of the MCP joint at each sampled position varies within the range of 5 to 15 Nmm, depicted in blue and orange, respectively in Fig.~\ref{fig: result ID h 1}(b). The measured torques applied by the external motor are shown in yellow and purple for the flexion and extension phases with standard deviation, respectively.  
As a result, the $ \myvec h(\myvec q, \myvec{\dot{q}}) $ torque of the MCP joint in this static phase is derived according to the Eq.~\eqref{eq: ID.3 joint viscoelasticity} and fit into a 2nd-order polynomial depicted as blue curves in Fig.~\ref{fig: result ID h 1}(c) while moving along flexion and extension direction.
The mean curve of two blue curves is considered as direction-independent elastic torque $\tau_k$, which reaches the maximal extension torque of 30 Nmm at 45$^{\circ}$. The error between the blue curves and the red curve is taken as the direction-dependent friction torque $\tau_f$ of the MCP joint, which achieved the maximal absolute value of 9 Nmm at 70$^{\circ}$.  



In the dynamic phase of the ID.3 experiment, the MCP joint is driven back and forth within a range of 10$^{\circ}$ to 65$^{\circ}$ at joint velocities of 0.5, 1, and 1.5 rad/s, as shown in Fig.~\ref{fig: result ID h visco 1}(a).
The joint viscoelasticity torques at different joint velocities are illustrated in Fig.~\ref{fig: result ID h visco 1}(b) against the joint angle. 
From the results, the joint velocity has few effects on the joint viscoelasticity torque, indicating that the damping or viscous friction torques of the MCP joint are close to zero. 

\subsection{ID.4 Tendon transmission friction}

\begin{figure}[tb]
    \centering
    \includegraphics[width=0.48\textwidth]{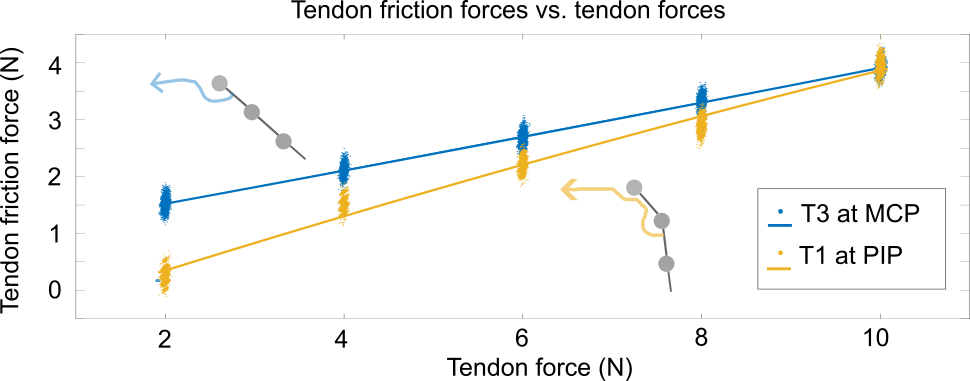}
    \caption{Examples of characteristic curves of tendon friction forces against the tendon forces. The X-axis is the applied tendon force at the actuation side. The Y-axis refers to the reductions of tendon force. The blue measurement points and associated fitted curve depict the friction force of T3 measured at the MCP joint. The yellow represents the friction force of T1 measured at the PIP joint.  }
    \label{fig: result ID ff}
    \vspace{-10pt}
\end{figure}

The reduction of tendon forces during the transmission system is captured by comparing the applied tendon force at the actuation side and the estimated force according to the model and measurement. \review{Fig.} \ref{fig: result ID ff} shows the tendon friction curves of two tendons, i.e. T1 and T3. 
The blue curve, is fit from the measurement data points by a 1st-order polynomial, reveals the force reduction of T3 through the MCP joint, which is fixed at 50$^{\circ}$ by the external torque sensor, as illustrated in the sketch. The tendon force of T3 increases from 4 to 10 N while the other tendons remain slack. With the model identified from the previous experiments, the tendon force reduction rises from 1.5 to 4 N. 
This indicates that the finger has 60\% tendon force transmission efficiency when T3 is in tension with 10 N. The yellow curve represents the tendon transmission friction of T1 through the MCP and PIP joints. The finger holds the pose of 25$^{\circ}$ at the MCP joint and 60$^{\circ}$ at the PIP joint. The PIP joint is connected to the external torque sensor. From the results, T1 has a smaller tendon friction than T3 when the tendon force is low. However, the slope of the T1 curve is higher than that of T3, indicating T1 has a larger friction coefficient along the tendon path. 
This might be explained by the finger design that T1 has an attachment with the edge of the tendon tunnel at the MCP joint, where the friction coefficient is larger than the surface attachment of T3.

\subsection{VD.2 Fingertip force estimation}

\begin{figure}[tb]
    \centering
    \includegraphics[width=0.48\textwidth]{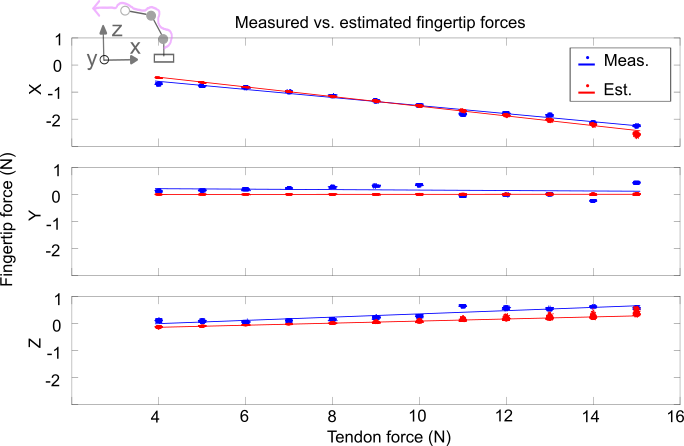}
    \caption{ The comparison of the measured and estimated fingertip force while pulling T2 (extensor) with increasing tension forces. The blue and red dots refer to the measured and estimated fingertip force, respectively, and are fitted into straight lines. }
    \label{fig: result VD fingertip}
    \vspace{-15pt}
\end{figure}

In this experiment, the finger is placed in a random configuration of 10$^{\circ}$, 40$^{\circ}$, and 35$^{\circ}$ at the MCP, PIP, and DIP joints, respectively, such that the finger pose can be maintained only by contact force with the thimble while pulling the T2.  
The finger pose and associated coordinates are visualised on the top left of Fig.~\ref{fig: result VD fingertip}.
In this pose, the x-axis is the dominant force direction of the contact. T2 is pre-tensioned with a tendon force of 4 N. Then, the tendon force was increased by 1 N each time step and eventually reached 15 N. The resultant contact force at the fingertip was recorded and illustrated in blue in Fig.~\ref{fig: result VD fingertip}. The red points represent the estimation of the contact force based on the identified model. The measured contact force along the x-axis increases from 0.7 N to 2.2 N in absolute value, while the estimated force rises from 0.45 N to 2.6 N. The estimated force along the y-axis remains 0 as the abduction joint is not considered in the model. The measured forces along the y- and z-axis vary within the range of 0.5 N, which might be due to the rotational error of the force sensor placement. The data points are fit with first-order polynomials and represented by lines. The coefficients for proportional terms of the measured and estimated force along the x-axis are -0.1485 and -0.1788, respectively.

\section{DISCUSSION AND CONCLUSIONS}

We formulated a general form of the dynamic model for an anthropomorphic tendon-driven robotic finger. The model included rigid body dynamics, nonlinear moment arms, passive viscoelasticity of joints, and tendon friction. The proposed methodology systematically identified and validated all model parameters through a sequence of experiments, drawing inspiration from existing identification approaches for cadaver hands \cite{an1983tendon}\cite{ francis2021moment} and robotic fingers \cite{valero2000quantification}, as well as novel experiments for acquiring joint viscoelasticity characteristics refined from previous studies \cite{deshpande2011contributions}.

Consequently, an identification test-bed contained 18 actuation modules along with associated peripheral sensors were developed.
Compared to the test-bed setups described by \cite{francis2021moment} and \cite{valero2000quantification}, our system incorporated a tendon force control algorithm that enabled precise control of tendon forces according to a desired trajectory. This was achieved through real-time tendon force sensing, rather than relying on constant load application. Moreover, an external motor equipped with a torque sensor provided the capability of manipulating the joint, instead of relying on tendon forces. In this case a tendon-driven anthropomorphic finger was designed based on the index finger of the Dexmart hand \cite{borghesan2010design} to validate the proposed methodology.

The tendon excursion method proposed in \cite{an1983tendon} was applied in ID.2 to determine moment arm curves at each joint by manipulating the joint individually and recording the corresponding tendon excursion. The results captured the non-linearity caused by the eccentric attachment surfaces. The identified coupling matrix was validated in VD.1 by the online estimation of the tendon excursion and comparing it to the measurement. 

Compared with the identification study \cite{tigue2019calibration}, in which the joint viscoelasticity identification was missing, we designed the static and dynamic phase of the ID.3 experiment to address the elastic and damping properties sequentially. 
In order to test the identification methodology, we incorporated a rubber band at the MCP joint since the original finger design of the Dexmart hand does not have any elastic element at joints \cite{borghesan2010design}.

In the ID.4 experiment, we considered a friction model to describe tendon friction. Compared to the study \cite{palli2011modeling}, in which the tendon friction model was identified in a standalone setup, we quantified the friction model parameters in-situ based on the identified dynamic model from ID.1 to ID.3. 

Due to the fact that the finger is a low-mass system, the inertia and Coriolis torques play a minor role in the system dynamics. Hence, we use the static fingertip force estimation as a validation scenario, which required an accurate modelling of all joint pose-related terms. However, the limitation of the current experimental setup with respect to the measurement was that the joint position acquisition and connection to the external torque sensor required the modification of the finger design, i.e. embedded encoders and salient shafts. This can be improved by replacing the joint encoder with the motion capture system, and designing a fixture component between the finger and torque sensor such that the test-bed can be used for identifying other tendon-driven fingers. 
Another limitation of this methodology is the identification of the joint torque caused by passive tendons since it is difficult to measure or estimate the passive tendon force without disassembling the finger during identification experiments. For example, the method of calculating the passive tendon force based on the tendon elongation and stiffness coefficient, proposed in \cite{borghesan2010design}, requires a more accurate joint position sensing based on the high stiffness of a Dyneema tendon.
\review{The abduction movement of the MCP joint is fixed in this study. The proposed identification method is not limited as long as the finger has a serial kinematic structure and each joint can be moved independently. Furthermore, material deformation and changes in properties are unavoidable in a 3D-printed finger after prolonged usage. However, since the experiments do not require high tendon forces or large force contacts, the entire process can be completed within a time range during which the material properties remain stable. }

Our study serves as a preliminary test of the methodology, which will be applied to identify a complete musculoskeletal hand model from a single cadaver hand to preserve data consistency.  Therefore, the proposed experiments were designed with the consideration of targeting the physiological hand properties. For instance, the joint viscoelasticity torque is the dominant property affecting human finger dynamics \cite{deshpande2011contributions}, and the fingertip force measurement is a prevalent way of validating musculoskeletal hand models \cite{mcfarland2022musculoskeletal}\cite{valero2000quantification}. 

Future work includes the further development of the test-bed to adapt to the cadaver hand studies, including the external motion capture system, a flexible supporting frame to fix the cadaver hand, and a fixture component for the external motor to access all the joints within the hand. Moreover, in a dynamic scenario, the finger is manipulated along a trajectory by tendon forces, which is used to validate the entire dynamic model through inverse dynamics. The cadaver hand study will be conducted following the proposed identification pipeline to identify a complete dynamic model of the human hand from one cadaver with all extrinsic muscles connected to the test-bed.







\printbibliography

@article{goislard2018importance,
  title={Importance of consistent datasets in musculoskeletal modelling: a study of the hand and wrist},
  author={Goislard De Monsabert, Benjamin and Edwards, Dafydd and Shah, Darshan and Kedgley, Angela},
  journal={Annals of biomedical engineering},
  volume={46},
  pages={71--85},
  year={2018},
  publisher={Springer}
}

@article{li2024fingertip,
  title={The Fingertip Manipulability Assessment of Tendon-driven Multi-fingered Hands},
  author={Li, Junnan and Ganguly, Amartya and Figueredo, Luis FC and Haddadin, Sami},
  journal={IEEE Robotics and Automation Letters},
  year={2024},
  publisher={IEEE}
}

@inproceedings{borghesan2010design,
  title={Design of tendon-driven robotic fingers: Modeling and control issues},
  author={Borghesan, Gianni and Palli, Gianluca and Melchiorri, Claudio},
  booktitle={2010 IEEE International conference on robotics and automation},
  pages={793--798},
  year={2010},
  organization={IEEE}
}

@article{palli2014dexmart,
  title={The DEXMART hand: Mechatronic design and experimental evaluation of synergy-based control for human-like grasping},
  author={Palli, Gianluca and Melchiorri, Claudio and Vassura, Gabriele and Scarcia, Umberto and Moriello, Lorenzo and Berselli, Giovanni and Cavallo, Alberto and De Maria, Giuseppe and Natale, Ciro and Pirozzi, Salvatore and others},
  journal={The International Journal of Robotics Research},
  volume={33},
  number={5},
  pages={799--824},
  year={2014},
  publisher={SAGE Publications Sage UK: London, England}
}

@article{grebenstein2012hand,
  title={The hand of the DLR hand arm system: Designed for interaction},
  author={Grebenstein, Markus and Chalon, Maxime and Friedl, Werner and Haddadin, Sami and Wimb{\"o}ck, Thomas and Hirzinger, Gerd and Siegwart, Roland},
  journal={The International Journal of Robotics Research},
  volume={31},
  number={13},
  pages={1531--1555},
  year={2012},
  publisher={SAGE Publications Sage UK: London, England}
}

@article{zhu2022anthropomorphic,
  title={An anthropomorphic robotic finger with innate human-finger-like biomechanical advantages Part I: Design, ligamentous joint, and extensor mechanism},
  author={Zhu, Yiming and Wei, Guowu and Ren, Lei and Luo, Zirong and Shang, Jianzhong},
  journal={IEEE Transactions on Robotics},
  volume={39},
  number={1},
  pages={485--504},
  year={2022},
  publisher={IEEE}
}

@inproceedings{xu2016design,
  title={Design of a highly biomimetic anthropomorphic robotic hand towards artificial limb regeneration},
  author={Xu, Zhe and Todorov, Emanuel},
  booktitle={2016 IEEE International Conference on Robotics and Automation (ICRA)},
  pages={3485--3492},
  year={2016},
  organization={IEEE}
}

@article{mcfarland2022musculoskeletal,
  title={A Musculoskeletal Model of the Hand and Wrist Capable of Simulating Functional Tasks},
  author={McFarland, Daniel C and Binder-Markey, Benjamin I and Nichols, Jennifer A and Wohlman, Sarah J and de Bruin, Marije and Murray, Wendy M},
  journal={IEEE Transactions on Biomedical Engineering},
  volume={70},
  number={5},
  pages={1424--1435},
  year={2022},
  publisher={IEEE}
}

@article{engelhardt2020new,
  title={A new musculoskeletal AnyBody™ detailed hand model},
  author={Engelhardt, Lucas and Melzner, Maximilian and Havelkova, Linda and Fiala, Pavel and Christen, Patrik and Dendorfer, Sebastian and Simon, Ulrich},
  journal={Computer methods in biomechanics and biomedical engineering},
  volume={24},
  number={7},
  pages={777--787},
  year={2020},
  publisher={Taylor \& Francis}
}

@article{ma2019validated,
  title={A validated combined musculotendon path and muscle-joint kinematics model for the human hand},
  author={Ma’touq, Jumana and Hu, Tingli and Haddadin, Sami},
  journal={Computer methods in biomechanics and biomedical engineering},
  volume={22},
  number={7},
  pages={727--739},
  year={2019},
  publisher={Taylor \& Francis}
}

@article{mirakhorlo2018musculoskeletal,
  title={A musculoskeletal model of the hand and wrist: model definition and evaluation},
  author={Mirakhorlo, M and Van Beek, N and Wesseling, M and Maas, H and Veeger, HEJ and Jonkers, I},
  journal={Computer methods in biomechanics and biomedical engineering},
  volume={21},
  number={9},
  pages={548--557},
  year={2018},
  publisher={Taylor \& Francis}
}

@article{barry2018development,
  title={Development of a dynamic index finger and thumb model to study impairment},
  author={Barry, Alexander J and Murray, Wendy M and Kamper, Derek G},
  journal={Journal of Biomechanics},
  volume={77},
  pages={206--210},
  year={2018},
  publisher={Elsevier}
}

@inproceedings{tigue2019calibration,
  title={Calibration and Validation of Dynamic Model for Simulating Robotic Finger Kinematics and Contact Forces},
  author={Tigue, James A and Mascaro, Stephen A},
  booktitle={Dynamic Systems and Control Conference},
  volume={59148},
  pages={V001T03A003},
  year={2019},
  organization={American Society of Mechanical Engineers}
}

@article{deshpande2010acquiring,
  title={Acquiring variable moment arms for index finger using a robotic testbed},
  author={Deshpande, Ashish D and Balasubramanian, Ravi and Ko, Jonathan and Matsuoka, Yoky},
  journal={IEEE Transactions on Biomedical Engineering},
  volume={57},
  number={8},
  pages={2034--2044},
  year={2010},
  publisher={IEEE}
}

@article{deshpande2011mechanisms,
  title={Mechanisms of the anatomically correct testbed hand},
  author={Deshpande, Ashish D and Xu, Zhe and Weghe, Michael J Vande and Brown, Benjamin H and Ko, Jonathan and Chang, Lillian Y and Wilkinson, David D and Bidic, Sean M and Matsuoka, Yoky},
  journal={IEEE/ASME Transactions on mechatronics},
  volume={18},
  number={1},
  pages={238--250},
  year={2011},
  publisher={IEEE}
}

@article{francis2021moment,
  title={The moment arms and leverage of the human finger muscles},
  author={Francis-Pester, Fraser W and Thomas, Richard and Sforzin, David and Ackland, David C},
  journal={Journal of Biomechanics},
  volume={116},
  pages={110180},
  year={2021},
  publisher={Elsevier}
}

@article{an1983tendon,
  title={Tendon excursion and moment arm of index finger muscles},
  author={An, Kai-Nan and Ueba, Y and Chao, EY and Cooney, WP and Linscheid, RL},
  journal={Journal of biomechanics},
  volume={16},
  number={6},
  pages={419--425},
  year={1983},
  publisher={Elsevier}
}

@article{mirakhorlo2016anatomical,
  title={Anatomical parameters for musculoskeletal modeling of the hand and wrist},
  author={Mirakhorlo, Mojtaba and Visser, Judith MA and Goislard de Monsabert, BAAX and Van der Helm, FCT and Maas, Huub and Veeger, HEJ},
  journal={International Biomechanics},
  volume={3},
  number={1},
  pages={40--49},
  year={2016},
  publisher={Taylor \& Francis}
}

@article{deshpande2011contributions,
  title={Contributions of intrinsic visco-elastic torques during planar index finger and wrist movements},
  author={Deshpande, Ashish D and Gialias, Nick and Matsuoka, Yoky},
  journal={IEEE Transactions on Biomedical Engineering},
  volume={59},
  number={2},
  pages={586--594},
  year={2011},
  publisher={IEEE}
}

@article{kuo2012muscle,
  title={Muscle-tendon units provide limited contributions to the passive stiffness of the index finger metacarpophalangeal joint},
  author={Kuo, Pei-Hsin and Deshpande, Ashish D},
  journal={Journal of biomechanics},
  volume={45},
  number={15},
  pages={2531--2538},
  year={2012},
  publisher={Elsevier}
}

@article{li2006robot,
  title={A robot-assisted study of intrinsic muscle regulation on proximal interphalangeal joint stiffness by varying metacarpophalangeal joint position},
  author={Li, Zong-Ming and Davis, Gregg and Gustafson, Norm P and Goitz, Robert J},
  journal={Journal of orthopaedic research},
  volume={24},
  number={3},
  pages={407--415},
  year={2006},
  publisher={Wiley Online Library}
}

@article{palli2011modeling,
  title={Modeling, identification, and control of tendon-based actuation systems},
  author={Palli, Gianluca and Borghesan, Gianni and Melchiorri, Claudio},
  journal={IEEE Transactions on Robotics},
  volume={28},
  number={2},
  pages={277--290},
  year={2011},
  publisher={IEEE}
}

@inproceedings{lange2021friction,
  title={Friction estimation for tendon-driven robotic hands},
  author={Lange, Friedrich and Pfanne, Martin and Steinmetz, Franz and Wolf, Sebastian and Stulp, Freek},
  booktitle={2021 IEEE International Conference on Robotics and Automation (ICRA)},
  pages={6505--6511},
  year={2021},
  organization={IEEE}
}

@article{valero2000quantification,
  title={Quantification of fingertip force reduction in the forefinger following simulated paralysis of extensor and intrinsic muscles},
  author={Valero-Cuevas, Francisco J and Towles, Joseph D and Hentz, Vincent R},
  journal={Journal of Biomechanics},
  volume={33},
  number={12},
  pages={1601--1609},
  year={2000},
  publisher={Elsevier}
}

@article{lu2018novel,
  title={A novel experimental design for the measurement of metacarpal bone loading and deformation and fingertip force},
  author={Lu, Szu-Ching and Vereecke, Evie E and Synek, Alexander and Pahr, Dieter H and Kivell, Tracy L},
  journal={PeerJ},
  volume={6},
  pages={e5480},
  year={2018},
  publisher={PeerJ Inc.}
}

\end{document}